%% file: main.tex
\documentclass[10pt,twocolumn,letterpaper]{article}

\usepackage[pagenumbers]{cvpr} 

\input{preamble}

\definecolor{cvprblue}{rgb}{0.21,0.49,0.74}
\usepackage[pagebackref,breaklinks,colorlinks,allcolors=cvprblue]{hyperref}

\def\paperID{31612} 
\def\confName{CVPR}
\def\confYear{2026}

\title{Video-Only ToM: Enhancing Theory of Mind in Multimodal Large Language Models}

\author{Siqi Liu, Xinyang Li, Bochao Zou$^\dagger$, Junbao Zhuo, Huimin Ma, Jiansheng Chen$^\dagger$\\
University of Science and Technology Beijing\\
{\tt\small \{liusq, lxyyy\}@xs.ustb.edu.cn, \{zoubochao, junbaozhuo, mhmpub, jschen\}@ustb.edu.cn}
}

\begin{document}
\maketitle
\input{sec/0_abstract}    
\input{sec/1_intro}
\input{sec/2_related}
\input{sec/3_method}
\input{sec/4_experiments}
\input{sec/5_conclusion}
\section*{Acknowledgments}
This work was supported by the National Natural Science Foundation of China (62376024, 62576032, U25B2073),  the National Science and Technology Major Project (2022ZD0117902), and the Fundamental Research Funds for the Central Universities (FRF-TP-22-043A1). We thank the anonymous reviewers for insightful discussions.
{
    \small
    \bibliographystyle{ieeenat_fullname}
    \bibliography{main}
}

\input{sec/X_suppl}

\end{document}

%% file: preamble.tex
\usepackage{multicol}
\usepackage{multirow}
\usepackage{longtable}
\usepackage[accsupp]{axessibility}  

%% file: sec/0_abstract.tex
\begin{abstract}
As large language models (LLMs) continue to advance, there is increasing interest in their ability to infer human mental states and demonstrate a human-like Theory of Mind (ToM). Most existing ToM evaluations, however, are centered on text-based inputs, while scenarios relying solely on visual information receive far less attention. This leaves a gap, since real-world human–AI interaction typically requires multimodal understanding. In addition, many current methods regard the model as a black box and rarely probe how its internal attention behaves in multiple-choice question answering (QA). The impact of LLM hallucinations on such tasks is also underexplored from an interpretability perspective. To address these issues, we introduce VisionToM, a vision-oriented intervention framework designed to strengthen task-aware reasoning. The core idea is to compute intervention vectors that align visual representations with the correct semantic targets, thereby steering the model's attention through different layers of visual features. This guidance reduces the model's reliance on spurious linguistic priors, leading to more reliable multimodal language model (MLLM) outputs and better QA performance. Experiments on the EgoToM benchmark—an egocentric, real-world video dataset for ToM with three multiple-choice QA settings—demonstrate that our method substantially improves the ToM abilities of MLLMs. Furthermore, results on an additional open-ended generation task show that VisionToM enables MLLMs to produce free-form explanations that more accurately capture agents' mental states, pushing machine–human collaboration toward greater alignment.
\begingroup
\renewcommand\thefootnote{}\footnote{Project Page: https://founce.github.io/VisionToM}\footnote{$^\dagger$ Corresponding authors.}
\addtocounter{footnote}{-1}
\endgroup
\end{abstract}

%% file: sec/1_intro.tex
\section{Introduction}
\begin{figure*}
  \centering
  \includegraphics[width=1.0\linewidth]{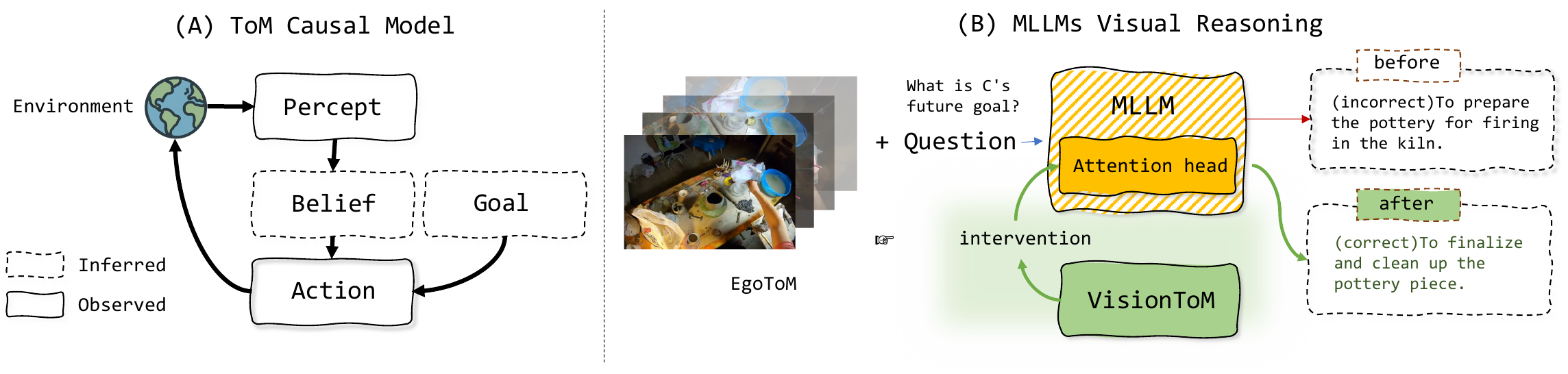}
  \caption{{
 (A). ToM Causal Model \cite{li2025egotom}
 (B). An overview of our method: MLLMs' visual reasoning with VisionToM intervention on the EgoToM benchmark. Given an egocentric video and a ToM question (e.g., ``What is C's future goal?"), a MLLM may produce an incorrect answer based on its default attention. VisionToM extracts representations from the MLLM for visual attention and ToM reasoning, identifies attention heads sensitive to visual input and task-specific reasoning, and performs targeted interventions on these heads. This process guides the model toward accurate, goal-consistent inferences aligned with ToM reasoning.}}
  \label{overview}
\end{figure*}

Theory of Mind (ToM) refers to the ability to impute mental states to self and others, including desires, beliefs, and intentions, in order to predict behavior \cite{baron1985does}. ToM is an essential component of human social intelligence, supporting complex interactions including communication, cooperation, empathy, and deception.

In humans, ToM typically develops gradually during early childhood through social experience. Over the past few decades, psychologists have developed a range of paradigms to study the development of ToM, such as the false belief task \cite{wimmer1983beliefs,baron1985does}, implicit inference paradigms \cite{onishi200515}, and eye-tracking techniques \cite{southgate2007action}. These paradigms inspire machine psychology, which compares AI and human mental-state reasoning on similar tasks \cite{rabinowitz2018machine}.

Recent studies offer mixed views on whether LLMs possess ToM abilities. While models like GPT-4 show human-like reasoning in some text-based tasks \cite{strachan2024testing}, these abilities are fragile and disrupted by minor input changes or added modalities \cite{van2024investigating}, suggesting reliance on surface patterns rather than interpretable psychological representations. Current evaluations remain limited to text inputs \cite{sap2022neural}, despite real-world ToM relying on multimodal, dynamic perception. Human social cognition unfolds over time in natural settings, indicating that first-person video may offer a more ecologically valid testbed for ToM reasoning \cite{chen2025through}.

Most multimodal ToM benchmarks rely on simulated environments—such as grid-worlds or controlled 3D scenes \cite{li2025black,kovavc2021socialai,chevalier2023minigrid,jin2024mmtom,shi2025muma}—which, despite offering experimental control, lack the perceptual richness of real-world settings. As a result, findings may not generalize to embodied agents operating in natural environments. In contrast, egocentric video provides more ecologically valid scenarios, requiring inference from partial, dynamic visual input. Moreover, multimodal large models (MLLMs) are prone to hallucination, generating ungrounded responses in ToM tasks. Some approaches have explored using interpretability techniques to enhance machine ToM capabilities, but these remain limited to the textual modality \cite{zhu2024language}.

To address these limitations, we propose integrating learnable intervention vectors into the model's attention layers, as illustrated in Figure~\ref{overview}(B). These vectors guide the model to attend to critical visual regions and features, thereby enhancing reasoning accuracy. Learned in the latent space, the intervention vectors are optimized to serve two primary objectives: enhancing visual attention and guiding ToM reasoning. Our method does not rely on handcrafted prompts or external linguistic annotations, and is compatible with arbitrary multi-class tasks, whereas traditional methods only handle binary ToM tasks. Since intervention vectors are computed once with the MLLM backbone frozen and reused at inference time, our method demonstrates strong task generalizability. We evaluate three core ToM reasoning tasks in the EgoToM benchmark—goal, belief, and action inference—and observe substantial performance gains (see Section~\ref{result}). More experiments in the Supplementary Material further verify transferability. These tasks align closely with the causal reasoning structure underlying cognitive models of ToM (see Section~\ref{ToM}).

In summary, our main contributions are: (1) We provide an interpretable analysis showing that MLLMs exhibit cross-task consistency in visual attention across multiple ToM tasks, while their internal ToM reasoning representations in hidden space diverge across tasks yet remain coherent within each task. This cross-task consistency of visual attention enables its targeted enhancement, and the intra-task uniformity of ToM reasoning internal representations allows VisionToM to probe task-specific ToM embeddings. (2) We introduce VisionToM, a lightweight backbone-frozen multimodal intervention framework that jointly enhances visual attention and ToM reasoning, requires no MLLM fine-tuning, handcrafted prompts, or external language annotations, and operates solely on raw video inputs without textual descriptions. (3) We concentrate on visual reasoning in ToM and on open-ended question answering performance. On the real-world ToM evaluation dataset EgoToM, our method significantly enhances MLLMs' ToM capabilities and demonstrates more accurate natural-language answers to open-ended questions.

%% file: sec/2_related.tex
\section{Related Works}

\subsection{Machine Theory of Mind} 
\label{ToM}
\subsubsection{Text-Based Machine ToM Evaluation.}
Text-based ToM reasoning has become a crucial direction for evaluating the social intelligence of LLMs. Early studies proposed several datasets to test models' abilities to infer beliefs and intentions from narrative texts. For example, the ToM Task dataset \cite{nematzadeh2018evaluating} offers short stories with targeted mental state questions, serving as a benchmark for assessing belief and desire inference. Similarly, the GLUCOSE dataset \cite{mostafazadeh2020glucose} provides narrations annotated with commonsense knowledge to support evaluation of causal and intentional reasoning capabilities. The Neural Theory-of-Mind (TOMI) dataset \cite{sap2022neural} further explores the boundaries of LLMs in attributing mental states, indicating that even state-of-the-art models face challenges in maintaining consistent and stable ToM reasoning. To enhance ToM abilities, researchers have explored prompt engineering strategies to guide models toward more effective mental state inference \cite{wilf2023think}. However, the current performance of LLMs remains questionable, often unstable and easily disrupted \cite{verma2024theory,kim2023fantom}.

\subsubsection{Theory of Mind Benchmarks for MLLMs.}
Multimodal datasets extend ToM evaluation into richer scenarios involving the fusion of video, image, and language inputs. These benchmarks typically require the model to observe video segments and answer reasoning questions related to characters' intentions, beliefs, or emotions. For instance, the Social-IQ dataset \cite{zadeh2019social} features YouTube interactions from real-life scenarios, using multiple-choice questions to assess social perception understanding. Other datasets, such as TVQA \cite{lei2018tvqa}, PororoQA \cite{kim2017deepstory}, and MovieQA \cite{tapaswi2016movieqa}, focus on narrative comprehension in visual media, containing elements of social reasoning but primarily targeting event understanding rather than explicit mental state attribution. Unlike VQA tasks—which involve answering factual or descriptive questions about an image or video—ToM evaluation of MLLMs aims to emulate human ToM capabilities, focusing on high-level cognitive reasoning such as inferring others' intentions, beliefs, and knowledge states. This emphasis on causal reasoning distinguishes ToM tasks from traditional VQA or vision-language alignment tasks and highlights the hallucination challenges faced by MLLMs during such reasoning. Some recent works have begun constructing specialized benchmarks targeting multimodal ToM reasoning. MMToM-QA \cite{jin2024mmtom} fills a gap by specifically evaluating mental state inference from multimodal inputs, though its scenarios are limited to single-agent behavior. Muma-ToM introduces a more complex framework for multi-agent, multimodal ToM evaluation, incorporating rich social contexts and behavioral trajectories \cite{shi2025muma}. GridToM \cite{li2025black} proposes a novel benchmark that incorporates diverse belief testing tasks and perceptual information from multiple perspectives. However, existing MLLM ToM benchmarks still largely rely on textual input. Models' performance drops significantly when evaluated under video-only conditions (e.g., EgoToM). Our proposed method, VisionToM, introduces a method to enhance ToM ability from a visually dominant perspective. By injecting intervention vectors into the model's internal representational space to influence attention mechanisms, VisionToM significantly boosts reasoning accuracy and social cognition in video-only settings.

\subsection{Hallucination Phenomena in MLLMs}
Hallucination is a widespread issue in both LLMs and MLLMs. It was initially extensively studied in the context of LLMs, referring to the generation of information that is inconsistent with factual knowledge or is unconstrained by the input context. The causes of such hallucinations are typically linked to the model's data, training, and inference processes \cite{huang2025survey}. With the rapid advancement of MLLMs, the hallucination problem becomes even more complex in visual-language tasks. These models integrate visual encoders with language models and are capable of handling complex tasks such as visual question answering (VQA), video QA, and instruction \cite{gao2023llama,li2023blip}. Vision-language models such as CLIP \cite{radford2021learning} and Flamingo \cite{alayrac2022flamingo} jointly model image and text, enabling stronger cross-modal representation learning and reasoning across visual and linguistic inputs. However, generative large multimodal models remain susceptible to hallucinations in image description, such as object-existence hallucinations in detailed captioning \cite{zhai2023halle}. Current mitigation strategies are mainly categorized into three types: (1) Data-level optimization, such as pretraining or fine-tuning on high-quality video-text pairs \cite{bai2023qwen,hu2023ciem,you2023ferret}. (2) Architectural enhancements, including the introduction of finer-grained modality alignment mechanisms such as Connection Module Enhancing \cite{chen2024internvl} and Alignment Training Optimization \cite{sun2024aligning,zhao2023beyond,gunjal2024detecting,jiang2024hallucination}. (3) Interpretability and post-processing techniques, which involve analyzing the model's behavior for explainability and correcting the outputs at inference time \cite{li2025black,zhou2023analyzing,li2023inference,chen2025ict,huo2024self,wu2024noiseboost,zhu2025ibd}.

Despite progress, current methods still face limitations when addressing complex reasoning tasks involving ToM. Our proposed approach leverages an explainability-driven post-processing approach, specifically probing and intervening in attention heads that are sensitive during visual and ToM reasoning processes in MLLMs to reduce hallucinations. Our method significantly improves performance on the EgoToM benchmark, including goal inference, belief reasoning, and action inference. Notably, our method is backbone-frozen and applicable to multi-class scenarios.

%% file: sec/3_method.tex
\section{VisionToM}
\label{method}

\subsection{Models}

\begin{figure*}
  \centering
  \includegraphics[width=1.0\linewidth]{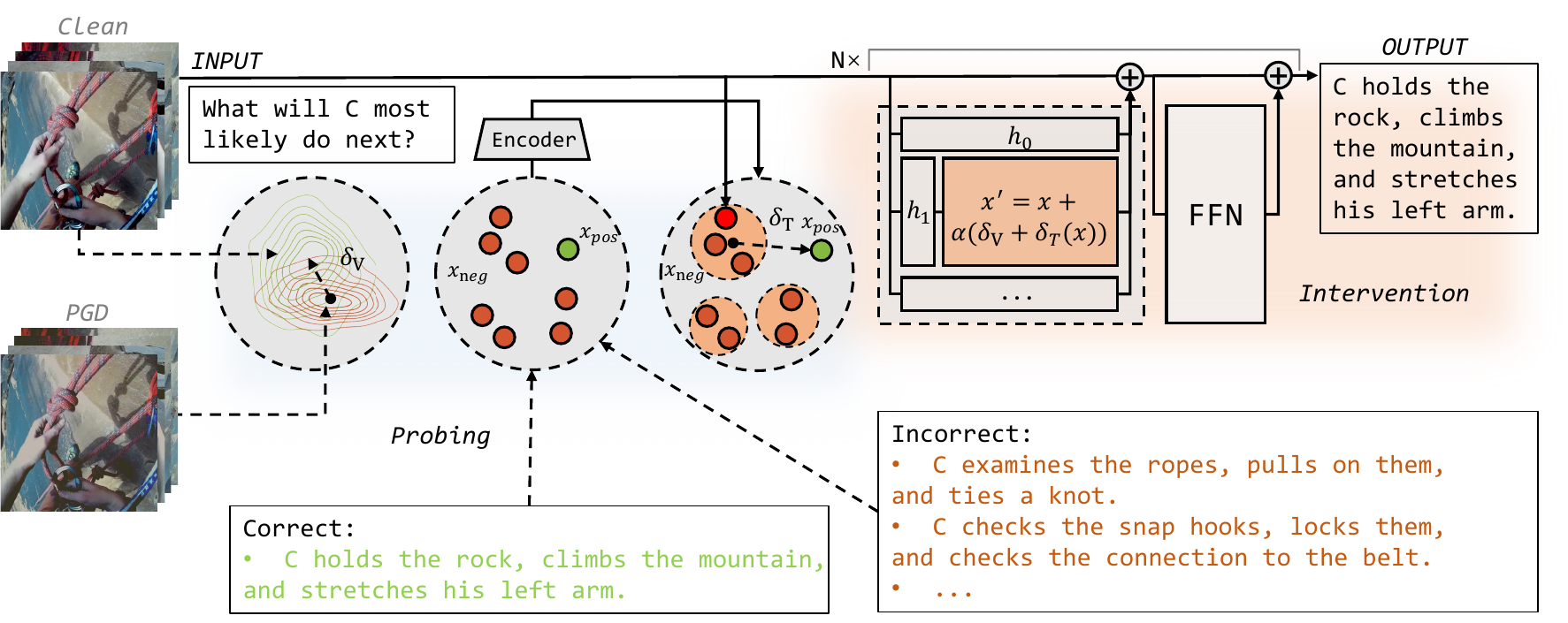}
  \caption{An overview of our method: we extract internal MLLMs representations along both visual and textual dimensions and identify attention heads that are sensitive to visual inputs and task reasoning. During inference, we then apply targeted interventions to these sensitive attention heads to enhance the MLLMs' truthfulness.}
  \label{mainmethod}
\end{figure*}

In the probing and intervention stages, we employ the LLaVA-Next-Video~\cite{zhang2024llavanextvideo} and Qwen2.5-VL~\cite{Qwen2.5-VL} models—both MLLMs were specifically designed for video understanding and generation tasks. To maintain clarity in our methodological exposition, this subsection focuses primarily on the LLaVA-Next-Video model; in Section~\ref{result} and the Supplementary Material, we further demonstrate that our approach is equally effective when applied to the Qwen2.5-VL model. Figure~\ref{mainmethod} outlines our method, specifically consisting of four parts in Sections~\ref{extractInternalRepresentations} -~\ref{intervention}.

\subsection{Extract Internal Representations}
\label{extractInternalRepresentations}

We begin by examining whether and how MLLMs represent different aspects of our tasks, encompassing both visual and textual dimensions. Since our task is primarily a vision-based reasoning problem—with inputs consisting solely of video and text-based questions, and no accompanying textual annotations—our goal is to decompose the visual ToM reasoning task into two components: visual representation and belief representation. We then decode these representations from the activations of attention heads.

Specifically, in our task, the visual input comprises video frames, while the textual input consists solely of the posed questions. We intentionally omit any textual annotations both to rigorously assess the visual reasoning capabilities of MLLMs and to minimize textual interference in the inference process. MLLMs begins by projecting its multimodal inputs into high-dimensional representations. Visual inputs \(V={v_1,v_2,\dots,v_m}\) and textual inputs \(X={x_1,x_2,\dots,x_n}\) are embedded separately, where \(m\) and \(n\) are their respective token counts. These embeddings are then concatenated into a single sequence \(T=concat(V, X) \in \mathbb{R}^{(m+n)\times DH}\) with \(D\) being the dimensionality per attention head and \(H\) the total number of heads. This combined input is fed into the Transformer of \(L\) layers to perform attention dot product. Within each layer, the sequence \(T_l\) is updated via multi‐head attention. The update from layer \(T_l\) to \(T_{l+1}\) is

\begin{equation}
\label{eq1}
T_{l+1} = T_{l} + \sum_{h=1}^{H} Attn^h_{l}(P_l^hT_{l}) \cdot W^o_{l},
\end{equation}

where \(Attn_l^h\) is the attention operation of head \(h\) at layer \(l\), \(P_l^h\in \mathbb{R}^{D\times DH}\) projects the layer's activations into the \(D\)-dimensional subspace for head \(h\), and \(W^o_{l}\in \mathbb{R}^{D\times DH}\) maps the aggregated head outputs back into the model's hidden space. Probing and intervention occur immediately after the attention computation and before the output projection.

Our approach decomposes the multimodal reasoning task into two fine-grained modules. The first module encourages the model to attend to visual inputs, thereby reducing its over-reliance on linguistic priors. The second module enhances ToM reasoning capabilities. In this section, we focus on the preliminary extraction of internal representations; the specific probing and intervention techniques are detailed in Sections~\ref{probing} and ~\ref{intervention}. Concretely, we construct positive and negative sample pairs tailored to each module.

\subsubsection{Visual Attention Enhancement.}
For the visual modality, unlike using random noise to guide attention \cite{chen2025ict}, we use the standard \(\ell^\infty\) bounded projected gradient descent (PGD) attack \cite{madry2018towards} on MLLM, making it outputs incorrect information, thereby generating adversarial examples. For the textual modality, we fix the question (e.g., for the ``Actions" task, the prompt ``What will C most likely do next?") and construct positive/negative pairs by varying adversarial noise. For each sample pair, we regard the representation of the final token as the fused multimodal embedding and extract the activations of attention heads that capture visual focus. There are \(H\times L\) attention heads over \(L\) layers for both positive and negative samples, denoted as \(X_V^{pos}={\{Attn^h_{l}(P_l^hT_{l}^{pos})\}}_{h=1,l=1}^{H,L}\) and \(X_V^{neg}={\{Attn^h_{l}(P_l^hT_{l}^{neg})\}}_{h=1,l=1}^{H,L}\). For all positive–negative activation pairs across $S$ samples, we compute an activation offset vector \({\{\delta_{V,l}^h\}}_{h=1,l=1}^{H,L}\), encouraging the model to focus more on visual information:

\begin{equation}
\label{eq2}
\{\delta_{V,l}^h\}=\frac{1}{S}\sum_{i=1}^{S} (X_{V,i,l}^{pos,h}-X_{V,i,l}^{neg,h}),
\end{equation}

\subsubsection{ToM Reasoning Guidance.}
In this phase, we fix the visual input to eliminate the influence from the images and vary only the textual inputs. We treat the correct answer as the positive sample and the set of incorrect answers as the negative samples. From these positive and negative samples, we again extract the attention-head activations associated with visual focus, denoted as \(X_T^{pos}={\{Attn^h_{l}(P_l^hT_{l}^{pos})\}}_{h=1,l=1}^{H,L}\) and \(X_T^{neg}={\{Attn^h_{l}(P_l^hT_{l}^{neg})\}}_{h=1,l=1}^{H,L}\). Because the semantic variation among negative samples leads to a non-uniform distribution of their attention representations in the hidden space, it is infeasible to derive a single offset vector from the negative-sample set toward the positive sample (see Figure \ref{mainmethod}). Consequently, we employ an encoder to separate the representations of positive and negative samples in Section~\ref{representationsSeparate}.

\subsection{Probing}
\label{probing}
\begin{figure*}[htbp]
  \centering
  \includegraphics[width=1.0\linewidth]{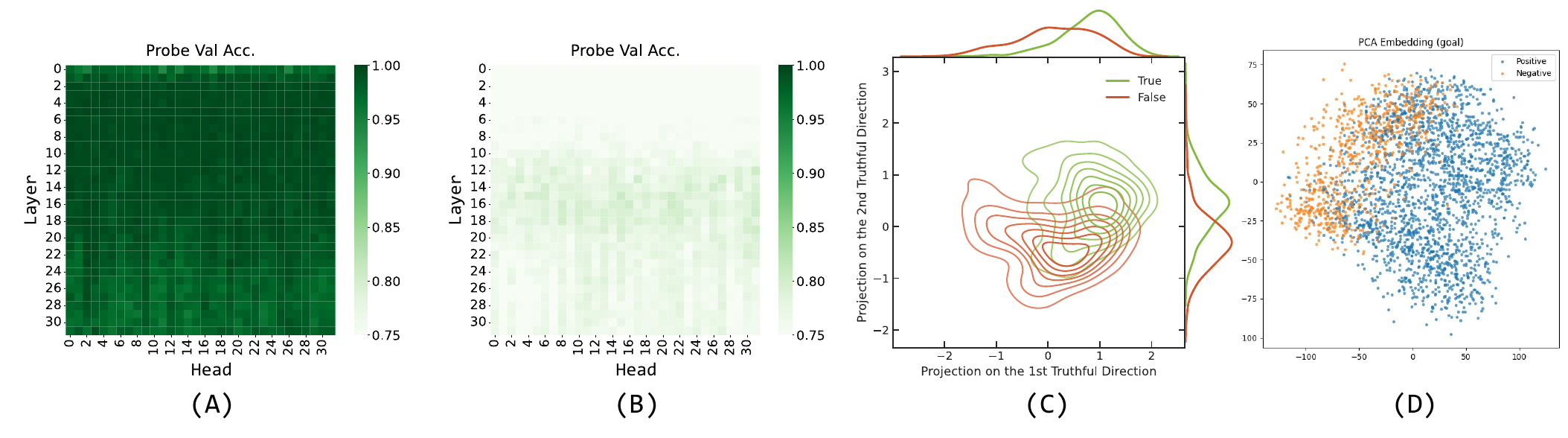}
  \caption{(A) Linear-probing accuracy for every head and layer of LLaVA-Next-Video on the visual-attention stage, incorporating internal representations from all three tasks. Darker green indicates higher accuracy, with 50\% marked as the chance baseline. (B) Linear-probing validation accuracy for every head and layer of LLaVA-Next-Video on the ToM-reasoning stage, incorporating internal representations from all three tasks. (C) Kernel density estimate (KDE) of LLaVA-Next-Video's visual-attention activations, projected onto the first two ``true" directions, showing the distributions for true (green) and false (orange) sample pairs. Marginal distributions are plotted along the top and right axes. (D) Principal component analysis (PCA) plot of LLaVA-Next-Video's internal representations in the ToM-reasoning stage.}
  \label{analysis-all}
\end{figure*}

Probing involves training a lightweight classifier on a network's activations to reveal how it encodes particular input or output characteristics \cite{kohn_whats_2015, gupta_distributional_2015,li2025black} as follows:

\begin{equation}
\label{eq3}
f_l^h = \frac{1}{1 + e^{-(\theta^Tx+b)}},
\end{equation}

where \(\theta\in \mathbb{R}^D\) and \(b\in \mathbb{R}\) represent the weight vector and bias, while \(f_l^h\) denotes a logistic sigmoid function for \((\theta^Tx+b)\), respectively. The parameters \(\theta\) and \(b\) are optimized by minimizing the cross-entropy loss. The probing experiments results are shown in Figure \ref{analysis-all}. More probing results are provided in the Supplementary Material.

For each attention head, we train a separate linear binary probe to fit the internal representations from each task. Specifically, we use a logistic regression model to predict the probability of the answer being true. The aim is to identify which heads are most task-sensitive—i.e., to determine which heads can distinguish \((<X_V^{pos}, X_V^{neg}>\) and which heads can distinguish \(<X_T^{pos}, X_T^{neg}>)\). Figures \ref{analysis-all}(A) and (B) show the validation accuracies of this probe on the two sets of positive/negative samples. Figure \ref{analysis-all}(A) indicates that many heads accurately capture the effect of visual noise, with these signals distributed across different layers and heads; by contrast, Figure \ref{analysis-all}(B) shows that heads sensitive to ToM reasoning tasks are concentrated in the middle layers, and their ability to discriminate positive from negative samples degrades as representations are forwarded. The results in the Supplementary Material further confirm this: for the three ToM tasks, the middle layers are most discriminative. Especially for the Goal task of achieving high precision in the baseline, the middle layers peak in sensitivity, indicating that the model is more sensitive to the ToM reasoning for the Goal task, thus yielding higher accuracy on QA.

To better understand belief encoding in attention-head activation space, we visualize the geometric structure of the visual activations for all tasks in Figure \ref{analysis-all}(C). Specifically, we apply principal component analysis to reduce the activation vectors to two dimensions and select the two orthogonal directions of greatest variance to separate ``true" from ``false" features. The projected geometry reveals partially overlapping yet distinct distributions; notably, the second principal direction still exhibits a unique spread, suggesting that the notions of ``true" and ``false" inhabit a subspace within the attention space rather than a single unified axis. We also perform PCA on the internal representations of the Goal task in Figure \ref{analysis-all}(D), finding that some negative samples overlap with other positive samples within the same task group, though the overall distribution remains cohesive—explaining why certain heads, despite lower probe accuracy, nonetheless retain discriminative power.

In summary, the probing results provide an interpretable demonstration that MLLMs exhibit cross-task consistency in visual attention across multiple ToM tasks, while their multimodal internal representations in hidden space diverge across tasks yet remain coherent within each task. Therefore, in Section \ref{intervention}, we directly intervene on the $K$ attention heads that are most sensitive across all tasks based on this visual-attention consistency introduced above. In Section \ref{representationsSeparate}, we go beyond the strategy adopted by GridToM \cite{li2025black}, which derives intervention directions from the coefficient vectors of binary logistic-regression classifiers. In contrast, we implement a finer-grained procedure: for each ToM reasoning task we first embed each positive/negative sample pair with the encoder, obtain the pair-wise separation direction in representation space, and then intervene on the same top-$K$ sensitive heads identified for that task.

\subsection{Seperating ToM Reasoning Representations}
\label{representationsSeparate}

Building upon the findings from Section~\ref{probing}, we adopt a more fine-grained strategy that addresses the heterogeneity among different sample representations within the same task. Recognizing that different types of reasoning failures require distinct directional interventions, we employ a clustering-based approach to extract prototypes from negative samples and utilize encoders to disentangle the semantic spaces between each prototype cluster and positive samples. The encoder learns to provide translation vectors from representations, moving prototype cluster embeddings toward their corresponding positive sample embeddings to achieve more stable alignment between representations.

We begin by analyzing the internal representation space of negative samples to identify prototype clusters corresponding to different types of reasoning failures. For each attention head $h$ identified as predictive during the probing phase, we collect all negative sample representations $\{x_{T,i}^{\mathrm{neg},h}\}_{i=1}^{N_h}$, where $N_h$ denotes the number of negative samples for attention head $h$. To adapt to the attention head representations, we employ multiple clustering quality metrics to automatically determine the optimal number of clusters: (1) Silhouette Analysis~\cite{ROUSSEEUW198753}, (2) Elbow Method~\cite{thorndike1953belongs}, and (3) Calinski-Harabasz Index~\cite{calinski1974dendrite}.

The optimal cluster number $k^*_h$ is jointly determined by the three aforementioned objectives. Specifically, we comprehensively consider the cluster numbers that optimize the three metrics. In case of ties, we prefer smaller cluster numbers to avoid over-segmentation. The cluster number is constrained within the range $k^*_h \in [2,15]$, ensuring each cluster contains at least 5 samples to guarantee statistical reliability.

After determining $k^*_h$ clusters for each attention head $h$, we train encoders for each attention head to learn transformation patterns from various prototype clusters. Let $C_{h,c}$ denote the set of negative samples assigned to cluster $c$ for attention head $h$. The objective loss function is defined as:

\[
L_{\mathrm{total}} = \sum_{h} \sum_{c=1}^{k^*_h} \frac{1}{|C_{h,c}|} \sum_{i \in C_{h,c}} \left\| \bigl(x_{T,i}^{\mathrm{neg},h} + \delta_{h,c,i}\bigr) - x_{T,i}^{\mathrm{pos},h} \right\|^2,
\]

where the outer summation traverses all predictive attention heads $h$, the inner summation iterates over all clusters $c$ for head $h$, and $k^*_h$ denotes the optimal cluster number for head $h$, $x_{T,i}^{\mathrm{neg},h}$ represents the $i$-th negative sample representation for head $h$, $x_{T,i}^{\mathrm{pos},h}$ denotes the corresponding positive sample representation, $\delta_{h,c,i} = f_{h,c}(x_{T,i}^{\mathrm{neg},h})$ is the correction vector output by the cluster-specific encoder network, $f_{h,c}$ represents the specialized encoder network for the $c$-th cluster of head $h$, and $|C_{h,c}|$ indicates the number of samples in cluster $c$.

This loss function ensures that the corrected negative samples $(x_{T,i}^{\mathrm{neg},h} + \delta_{h,c,i})$ approximate their corresponding positive sample representations as closely as possible. This design allows each cluster-specific network to focus on correcting its corresponding type of reasoning failure while maintaining intra-cluster consistency.

During the intervention inference phase, the framework identifies the nearest cluster center using Euclidean distance, then employs the corresponding directional network for intervention. This framework ensures that each intervention targets the specific type of reasoning failure exhibited by the input, thereby achieving more precise and effective corrections, yielding interpretable intervention directions that enhance downstream QA performance.

\subsection{Intervention}
\label{intervention}

Despite the probing results demonstrating that MLLMs possess internal representations, we further seek to validate the practical effectiveness of these classifier-derived representation directions by intervening on attention heads. We derive the intervention direction \(\Delta\) from visual-attention and ToM-reasoning textual-representation separation as follows:

\begin{equation}
\label{eq4}
\Delta=\delta_{V,l}^h+ \delta_{T,l}^h\ .
\end{equation}

Then, we select the top \(K\) most sensitive attention heads identified during probing, separately for visual attention and for ToM reasoning, which are most attuned to distinctions between ``true" and ``false" representations. For the visual attention part, due to the existence of cross-task consistency, we maintain a common sensitive attention head, whereas for the ToM reasoning part we use separate sensitive attention heads for each task. During the MLLM inference phase, we apply interventions on these chosen heads immediately after the multi-head attention computation but before the projection back to the output layer, computed as follows:

\begin{equation}
\label{eq5}
T_{l+1} = T_{l} + \sum_{h=1}^{H} (Attn^h_{l}(P_l^hT_{l})+\alpha \times \Delta) \cdot W^o_{l},
\end{equation}

where \(\alpha\) is a scalar of intervention strength.

%% file: sec/4_experiments.tex
\section{Experiments}
\label{experiments}

\subsection{Baselines}

We conduct our experiments on EgoToM \cite{li2025egotom}, a new benchmark consisting of egocentric videos for real-world ToM evaluation. Unlike traditional action-recognition and VQA datasets, EgoToM specifically benchmarks agents' ToM abilities. Each instance in the dataset is paired with carefully curated question-answer pairs, including goal inference, belief reasoning, and action inference. The dataset covers diverse scenarios, capturing rich social interactions that challenge both perception and high-level reasoning, making it a suitable benchmark for evaluating embodied and cognitively grounded AI models.

In our experiments, we focus on video-only complete contextual information input (fullcontext in the EgoToM). In addition to the multiple models and two modalities provided in the EgoToM benchmark, we also tested several newer MLLMs, including both closed-source and open-source models. We then selected two open-source models to evaluate the effectiveness of our approach: LLaVA-Next-Video~\cite{zhang2024llavanextvideo}, specifically designed for video-based instruction following and question answering, and Qwen2.5-VL~\cite{Qwen2.5-VL}, a large-scale multilingual vision-language model that excels in cross-lingual and multimodal understanding. More models are detail in the Supplementary Material.

\subsection{Settings}

\subsubsection{Visual Attention Enhancement.} To obtain informative negative samples that expose attention failures, we generate adversarially perturbed frames and use them to approximate a correction direction in the latent space. We use an \(\ell^\infty\)-bounded PGD attack that maximizes the cross-entropy loss on the ground-truth answer by backpropagating to the normalized 24-frame video tensor before vision encoding. We parameterize the PGD attack with a perturbation bound of $\epsilon = \frac{16}{255}$, a step size of $\frac{1}{255}$, and a total of $T = 300$ iterations. To benchmark the effectiveness of this attack, we also ran a set of experiments using random Gaussian noise, where the noise standard deviation $\sigma \in [50,\,80]$, and all perturbed pixel values were clipped to remain within the valid range.
\subsubsection{ToM Reasoning Guidance.} The encoder consists of two linear layers—each followed by a GELU activation and layer normalization—and, for each attention head, maps dimensions 128 $\rightarrow$ 256 $\rightarrow$ 128. During training, we use the Adam optimizer with a learning rate of \(1\times10^{-3}\), and each cluster learns a dedicated directional correction tailored to its corresponding reasoning failure patterns. The probe and encoder are trained once while the MLLM backbone stays frozen.
\subsubsection{Intervention.} We set our best performance configuration as edit heads $k=64$, intervention strength $\alpha=1.0$, with the performance impact of parameters analyzed in Section~\ref{result}. During the model inference process, we only input the video, question, and options, without providing any additional prompts. The model settings remain consistent with previous works~\cite{li2025black,zhu2024language}, following a zero-temperature zero-shot setting.

\subsection{Evaluation Protocol}
For each EgoToM task, we train the probe and encoder on a 30\% calibration split, then keep the resulting intervention vectors fixed for inference on a disjoint 70\% evaluation split without using its labels or answers. We follow the official evaluation and report Top-1 accuracy on the three subtasks: goal inference, belief reasoning, and action inference. Each video-question pair is associated with multiple candidate answers, and the model is required to output one final choice; for open-ended generation we additionally score the responses with the TruthfulQA-style \cite{lin-etal-2022-truthfulqa} rubric described in Section~\ref{result}. We also ensure that the intervention strength and edited heads are fixed across models when comparing different backbones, so that improvements can be attributed to the proposed VisionToM procedure rather than model-specific tuning. On the hardware reported in the Supplementary Material, the one-time calibration stage takes approximately 0.2 hours for probe training and 1 hour for encoder training.

\subsection{Results}
\label{result}

\setlength{\tabcolsep}{1mm}
\begin{table}[t]
  \centering
  \small
  \caption{Comparison of the effectiveness of random Gaussian noise attacks and PGD attack methods on LLaVA-Next-Video.}
  \label{randomVSPGD-table}
  \begin{tabular}{cccc|ccc}
    \toprule
    \multirow{2}{*}{Method} & \multicolumn{3}{c|}{Acc (\%) $\uparrow$} & \multicolumn{3}{c}{Acc after \(+\alpha\Delta\)(\%) $\uparrow$}\\
    \cmidrule(r){2-7}
     & Goal & Belief & Actions & Goal & Belief & Actions\\
    \midrule
    Baseline & 61.5 & 38.9 & 24.0 & \multicolumn{3}{c}{-}\\ \cmidrule{1-7}
    Random & 47.0 & 35.3 & 25.1 & 70.4 & 39.2 & 24.9\\
    PGD & 29.1 & 22.2 & 16.4 & \textbf{74.5} & \textbf{45.3} & \textbf{29.7}\\
    \bottomrule
  \end{tabular}
\end{table}

First, we compared the effects of random Gaussian noise and PGD attacks on LLaVA-Next-Video model using the EgoToM dataset, as shown in Table~\ref{randomVSPGD-table}. The results indicate that PGD attacks are more effective than random noise in reducing the model's reasoning accuracy for each task. Furthermore, we used two types of images as negative samples for visual attention enhancement. The results show that adversarial samples generated by PGD attacks provide more accurate guidance directions compared to random noise. In other words, feature direction estimation guided by adversarial samples is more valuable for improving ToM.

\setlength{\tabcolsep}{1mm}
\begin{table}[t]
  \centering
  \small
  \caption{Performance comparison of VisionToM with human baselines and multiple MLLMs' baselines on ToM tasks in the EgoToM benchmark.}
  \label{result-table}
  \begin{tabular}{cccccc}
    \toprule
    \multirow{2}{*}{Method} & \multirow{2}{*}{Setting} & \multirow{2}{*}{Nframe} & \multicolumn{3}{c}{Accuracy (\%) $\uparrow$} \\
    \cmidrule(r){4-6}
     & & & Goal & Belief & Actions \\
    \midrule
    \multirow{2}{*}{Humans} & \multirow{2}{*}{Baseline} & last 30s & 88 & 72 & 78\\
     &  & last 5s & 89 & 71 & 77\\ \cmidrule{1-6}
    GPT-4-Turbo & \multirow{7}{*}{Baseline} & 20 & 83 & 45 & 42\\
    Video-Llama2-72B &  & 8 & 85 & 46 & 40\\
    CogVLM2 &  & 24 & 73 & 39 & 36\\
    GPT-4o &  & 24 & 68.7 & 20.4 & 22.6\\
    Gemini-2.5-Flash &  & 24 & 86.0 & 46.7 & 40.1\\ \cmidrule{1-6}
     \multirow{6}{*}{\shortstack{LLaVA-Next-\\Video-7B}} & Baseline & \multirow{6}{*}{24} & 61.5 & 38.9 & 24.0\\
     & w/o \(\delta_{T,l}^h\) &  & 73.2 & 39.2 & 25.3\\
     & w/o \(\delta_{V,l}^h\) &  & 72.6 & \textbf{45.3} & 29.0\\
     & Rnd-\(\Delta\) &  & 62.1 & 39.2 & 25.4\\
     & \(-\alpha\Delta\) &  & 50.6 & 20.6 & 10.1\\
     & \(+\alpha\Delta\) &  & \textbf{74.5} & \textbf{45.3} & \textbf{29.7}\\ \cmidrule{1-6}
     \multirow{6}{*}{\shortstack{Qwen2.5-\\VL-7B}} & Baseline & \multirow{6}{*}{24} & 86.9 & 35.6 & 31.1\\
     & w/o \(\delta_{T,l}^h\) &  & \textbf{88.9} & 35.6 & 33.2\\
     & w/o \(\delta_{V,l}^h\) &  & 88.4 & 40.6 & 37.5\\
     & Rnd-\(\Delta\) &  & 87.1 & 36.0 & 31.9\\
     & \(-\alpha\Delta\) &  & 60.3 & 24.3 & 14.3\\
     & \(+\alpha\Delta\) &  & \textbf{88.9} & \textbf{42.0} & \textbf{37.6}\\
    \bottomrule
  \end{tabular}
\end{table}

Table~\ref{result-table} shows the most representative baseline results on the EgoToM dataset, baselines for additional newer MLLMs including LLaVA-Next-Video and Qwen2.5-VL models, the performance improvements achieved by our method, and ablation experiment results. First, all MLLMs, including LLaVA-Next-Video and Qwen2.5-VL, perform poorly on Belief and Actions tasks, with a significant gap compared to human baselines of 72\% and 78\%, indicating that ToM reasoning tasks remain challenging for MLLMs. Some models approach human baselines on the Goal task, for example, Gemini-2.5-Flash and Qwen2.5-VL achieve 86.0\% and 86.9\% accuracy on the Goal task, respectively.

Our method brings significant improvements: applying both visual attention and ToM reasoning interventions (\(+\alpha\)) can produce substantial gains, such as the LLaVA-Next-Video model improving by 13.0\%, 6.4\%, and 5.7\% on the three tasks respectively, and Qwen2.5-VL's results on the Goal task even match human level. In comparison, we selected the best results using random \(\Delta\) with the same settings, which showed no significant changes. To more clearly demonstrate the effects of visual attention and ToM reasoning interventions, we conducted ablation experiments, as shown in Table~\ref{result-table}. The results indicate that using only visual attention intervention or only ToM reasoning intervention can both effectively improve accuracy, and the accuracy is less than the result when both are applied simultaneously. This shows that the hidden layer features detected by visual attention and ToM reasoning respectively guide the visual and reasoning parts, operating along their respective effective directions, and the action directions are consistent with theoretical expectations, rather than canceling each other out. Notably, visual attention interventions are particularly effective for the Goal task; in contrast, the more challenging Belief and Action tasks in the ToM causal model (Figure \ref{overview}A) require reasoning interventions to produce effects. Additional large-model and cross-dataset results are deferred to the Supplementary Material. The hyperparameter impact analysis of the number of editing heads \(K\) and intervention strength \(\alpha\) is in the Supplementary Material, and the results demonstrate the effectiveness of our method.

\setlength{\tabcolsep}{1mm}
\begin{table}[t]
  \centering
  \small
  \caption{Comparison of VisionToM's open-ended generation test performance on the EgoToM dataset.}
  \label{open-ended-qa-table}
  \begin{tabular}{ccccc}
    \toprule
    \multirow{2}{*}{Method} & \multirow{2}{*}{Task} & \multicolumn{3}{c}{Baseline $/$ +$\alpha\Delta$} \\
    \cmidrule{3-5}
              &        & True (\%) $\uparrow$ & Info (\%) $\uparrow$ & True $\land$ Info (\%) $\uparrow$ \\
    \midrule
    \multirow{3}{*}{\shortstack{Gemini-2.5-\\Flash}}
              & Goal   & 75.5 & 35.3 & 20.2 \\
              & Belief & 28.1 & 99.7 & 28.1 \\
              & Action & 21.8 & 100.0 & 21.8 \\
    \midrule
    \multirow{3}{*}{\shortstack{LLaVA-Next-\\Video-7B}}
              & Goal   & 8.5 $/$ 27.3 & 100.0 $/$ 99.9 & 8.5 $/$ 27.2 \\
              & Belief & 19.5 $/$ 32.9 & 99.7 $/$ 99.8 & 19.2 $/$ 30.8 \\
              & Action & 14.4 $/$ 25.9 & 99.7 $/$ 99.5 & 14.4 $/$ 25.8 \\
    \midrule
    \multirow{3}{*}{\shortstack{Qwen2.5-\\VL-7B}}
              & Goal   & 76.1 $/$ 78.2 & 14.2 $/$ 45.9 & 9.4 $/$ 35.4 \\
              & Belief & 29.0 $/$ 33.8 & 92.5 $/$ 90.6 & 23.7 $/$ 27.9 \\
              & Action & 19.2 $/$ 24.0 & 98.9 $/$ 95.7 & 18.6 $/$ 22.7 \\
    \bottomrule
  \end{tabular}
\end{table}

We also tested the open-ended generation capabilities of models after VisionToM intervention. Specifically, we adopt the TruthfulQA \cite{lin-etal-2022-truthfulqa} rubric: true measures whether every factual claim is correct, info measures whether the answer provides substantive information, and true $\land$ info is the proportion of answers that are both correct and informative. Two independent DeepSeek-R1 models act as judges, and a judgment is accepted only when the two models agree. Manual verification with three volunteers yields human--LLM agreement rates of 96.2\% for the ``true'' label and 93.5\% for the ``info'' label. The resulting scores are reported in Table \ref{open-ended-qa-table}. The specific open-ended generation results and judge details are shown in the Supplementary Material. The results indicate that our method is also helpful for improving open-ended generation results, showing that our method can extend from QA tasks to natural language generation tasks.

%% file: sec/5_conclusion.tex
\section{Conclusion}

In this paper, we propose VisionToM, a framework that relies solely on visual input without depending on additional information supplements, enhancing the truthfulness of MLLMs through visual attention enhancement and ToM reasoning guidance. The results demonstrate that VisionToM significantly enhances the ToM capabilities of MLLMs and exhibits more accurate natural language answers on open-ended generation tasks. We believe that VisionToM can bring stronger psychological attribution capabilities to MLLMs, enabling more trustworthy human-AI interactions in cognitively demanding social environments. By explicitly aligning visual evidence with task-specific mental-state inference, VisionToM also offers a new perspective for connecting interpretability methods with performance-oriented multimodal reasoning. We expect this work to facilitate richer socially aware agents that operate robustly in egocentric, dynamic environments.

%% file: sec/X_suppl.tex
\clearpage
\setcounter{page}{1}
\maketitlesupplementary

\section{MLLMs with EgoToM baseline}
We provide a complete introduction here to all the multimodal large language models (MLLMs) that we compared or used in the EgoToM benchmark. Specifically, the EgoToM benchmark includes human baselines, LLMs baselines, and MLLMs baselines. These models represent the state-of-the-art in text and vision-language processing. In our work, we focus on the performance of MLLMs, selecting representative open-source and closed-source models from the EgoToM benchmark, including GPT-4-Turbo~\cite{openai2024gpt4technicalreport}, Video-Llama2-72B~\cite{damonlpsg2024videollama2}, and CogVLM2~\cite{hong2024cogvlm2}. Additionally, we also added two widely-used closed-source models: GPT-4o~\cite{openai2024gpt4ocard} and Gemini-2.5-Flash~\cite{comanici2025gemini25pushingfrontier}, as well as open-source models LLaVA-Next-Video-7B~\cite{zhang2024llavanextvideo}, Qwen2.5-VL-7B~\cite{Qwen2.5-VL}, and the reasoning model GLM-4.1V-9B-Thinking~\cite{glmvteam2025glm41vthinkingversatilemultimodalreasoning}. Furthermore, we selected LLaVA-Next-Video-7B and Qwen2.5-VL-7B models as the base models for VisionToM, and through our method, both models achieved better performance.

We conducted quantification based on the metric charts in the EgoToM~\cite{li2025egotom} benchmark and list the complete data baseline here in Table~\ref{result-table-full}, along with additional experiments we added.

\begin{longtable}{lcccccc}
\caption{Full Baseline and results} \label{result-table-full} \\
\toprule
\multirow{2}{*}{Method} & \multirow{2}{*}{Setting} & \multirow{2}{*}{Context} & \multirow{2}{*}{Nframe} & \multicolumn{3}{c}{Accuracy} \\
\cmidrule(r){5-7}
& & & & Goal & Belief & Actions \\
\midrule
\endfirsthead

\toprule
\multirow{2}{*}{Method} & \multirow{2}{*}{Setting} & \multirow{2}{*}{Context} & \multirow{2}{*}{Nframe} & \multicolumn{3}{c}{Accuracy} \\
\cmidrule(r){5-7}
& & & & Goal & Belief & Actions \\
\midrule
\endhead

\midrule
\multicolumn{7}{r}{Continued on next page} \\
\midrule
\endfoot

\bottomrule
\endlastfoot

\multirow{2}{*}{Humans} & \multirow{2}{*}{Video} & last 30sec & \multirow{2}{*}{-} & 0.88 & 0.72 & 0.78\\
 &  & last 5sec & & 0.89 & 0.71 & 0.77\\ \cmidrule(r){1-7}
\multirow{5}{*}{Llama3.1-405b-instruct} & \multirow{5}{*}{Text} & full context & \multirow{5}{*}{-} & 0.82 & 0.44 & 0.48\\
 &  & last 30sec &  & 0.80 & 0.46 & 0.46\\
 &  & last 5ec &  & 0.62 & 0.45 & 0.43\\
 &  & last action &  & 0.58 & 0.40 & 0.38\\
 &  & no context &  & 0.20 & 0.30 & 0.15\\ \cmidrule(r){1-7}
\multirow{5}{*}{Llama3.1-70b-instruct} & \multirow{5}{*}{Text} & full context & \multirow{5}{*}{-} & 0.80 & 0.34 & 0.47\\
 &  & last 30sec &  & 0.80 & 0.42 & 0.45\\
 &  & last 5ec &  & 0.65 & 0.41 & 0.42\\
 &  & last action &  & 0.60 & 0.36 & 0.38\\
 &  & no context &  & 0.28 & 0.25 & 0.18\\ \cmidrule(r){1-7}
\multirow{5}{*}{Llama3.1-8b-instruct} & \multirow{5}{*}{Text} & full context & \multirow{5}{*}{-} & 0.80 & 0.40 & 0.36\\
 &  & last 30sec &  & 0.78 & 0.42 & 0.38\\
 &  & last 5ec &  & 0.65 & 0.41 & 0.40\\
 &  & last action &  & 0.67 & 0.39 & 0.34\\
 &  & no context &  & 0.35 & 0.36 & 0.22\\ \cmidrule(r){1-7}
\multirow{5}{*}{GPT-4-Turbo} & \multirow{5}{*}{Video} & full context & \multirow{5}{*}{20} & 0.83 & 0.45 & 0.42 \\
 &  & last 30sec &  & 0.87 & 0.53 & 0.44 \\
 &  & last 5sec &  & 0.85 & 0.51 & 0.47 \\
 &  & last action &  & 0.78 & 0.50 & 0.41 \\
 &  & no context  &  & 0.15 & 0.18 & 0.06 \\ \cmidrule(r){1-7}
\multirow{2}{*}{GPT-4-Turbo} & \multirow{2}{*}{Text} & full context & \multirow{2}{*}{-} & 0.85 & 0.47 & 0.44 \\
 &  & last 30sec &  & 0.82 & 0.48 & 0.45 \\
\multirow{3}{*}{GPT-4-Turbo} & \multirow{3}{*}{Text} & last 5sec & \multirow{3}{*}{-} & 0.68 & 0.44 & 0.36 \\
 &  & last action &  & 0.60 & 0.34 & 0.32 \\
 &  & no context &  & 0.15 & 0.18 & 0.06 \\ \cmidrule(r){1-7}
\multirow{5}{*}{GPT-4} & \multirow{5}{*}{Text} & full context  & \multirow{5}{*}{-} & 0.86 & 0.46 & 0.47 \\
 &  & last 30sec &  & 0.82 & 0.48 & 0.43 \\
 &  & last 5sec &  & 0.70 & 0.42 & 0.41 \\
 &  & last action &  & 0.61 & 0.40 & 0.38 \\
 &  & no context &  & 0.20 & 0.28 & 0.18 \\ \cmidrule(r){1-7}
\multirow{5}{*}{GPT-3.5-Turbo} & \multirow{5}{*}{Text} & full context & \multirow{5}{*}{-} & 0.70 & 0.29 & 0.23\\
 &  & last 30sec &  & 0.70 & 0.32 & 0.21\\
 &  & last 5ec &  & 0.65 & 0.34 & 0.22\\
 &  & last action &  & 0.58 & 0.30 & 0.21\\
 &  & no context &  & 0.15 & 0.23 & 0.15\\ \cmidrule(r){1-7}
\multirow{5}{*}{VideoLLaMA2-72B} & \multirow{5}{*}{Video} & full context & \multirow{5}{*}{8} & 0.85 & 0.46 & 0.40\\
 &  & last 30sec &  & 0.86 & 0.48 & 0.42\\
 &  & last 5ec &  & 0.85 & 0.50 & 0.45\\
 &  & last action &  & 0.83 & 0.54 & 0.47\\
 &  & no context &  & 0.21 & 0.30 & 0.14\\ \cmidrule(r){1-7}
\multirow{5}{*}{VideoLLaMA2-7B-16F} & \multirow{5}{*}{Video} & full context & \multirow{5}{*}{16} & 0.67 & 0.33 & 0.30\\
 &  & last 30sec &  & 0.71 & 0.34 & 0.32\\
 &  & last 5ec &  & 0.73 & 0.41 & 0.34\\
 &  & last action &  & 0.66 & 0.39 & 0.36\\
 &  & no context &  & 0.32 & 0.25 & 0.19\\ \cmidrule(r){1-7}
\multirow{5}{*}{VideoLLaMA2-7B} & \multirow{5}{*}{Video} & full context & \multirow{5}{*}{8} & 0.79 & 0.41 & 0.31\\
 &  & last 30sec &  & 0.75 & 0.42 & 0.32\\
 &  & last 5ec &  & 0.75 & 0.40 & 0.40\\
 &  & last action &  & 0.52 & 0.33 & 0.35\\
 &  & no context &  & 0.32 & 0.28 & 0.21\\ \cmidrule(r){1-7}
\multirow{5}{*}{CogVLM2} & \multirow{5}{*}{Video} & full context & \multirow{5}{*}{24} & 0.73 & 0.39 & 0.36\\
 &  & last 30sec &  & 0.75 & 0.40 & 0.38\\
 &  & last 5ec &  & 0.77 & 0.42 & 0.41\\
 &  & last action &  & 0.53 & 0.34 & 0.32\\
 &  & no context &  & 0.21 & 0.29 & 0.30\\ \cmidrule(r){1-7}
GPT-4o & \multirow{5}{*}{Video} & \multirow{5}{*}{full context} & \multirow{5}{*}{24} & 0.69 & 0.20 & 0.23\\
Gemini-2.5-Flash &  &  &  & 0.86 & 0.47 & 0.40\\
GLM-4.1V-9B-Thinking &  &  &  & 0.80 & 0.31 & 0.26\\
LLaVA-Next-Video-7B &  &  &  & 0.62 & 0.39 & 0.24\\
Qwen2.5-VL-7B &  &  &  & 0.87 & 0.36 & 0.31\\
\end{longtable}

\newpage
\section{Additional Generalization Results}

We report two additional experiments promised in the rebuttal: scaling VisionToM to a stronger backbone and transferring the learned directions to a second video-only ToM benchmark.

\subsection{Large-Backbone Results}

\begin{table}[h]
  \centering
  \footnotesize
  \setlength{\tabcolsep}{0.6mm}
  \renewcommand{\arraystretch}{0.90}
  \caption{Additional experiments with a larger MLLM backbone. VisionToM continues to improve ToM reasoning when scaled to Qwen2.5-VL-72B.}
  \label{table-large-model}
  \begin{tabular}{@{}cccccc@{}}
    \toprule
    \multirow{2}{*}{Method} & \multirow{2}{*}{Setting} & \multirow{2}{*}{Nframe} & \multicolumn{3}{c}{Accuracy (\%) $\uparrow$} \\
    \cmidrule(r){4-6}
     & & & Goal & Belief & Actions \\
    \midrule
    {\scriptsize Gemini-3-Flash-Preview} & \multirow{2}{*}{Baseline} & \multirow{2}{*}{24} & \textbf{91.5} & 51.3 & 55.6\\
    {\scriptsize Qwen3-VL-235B-A22B-Instruct} &  &  & 90.2 & 51.3 & 44.4\\ \cmidrule{1-6}
     \multirow{6}{*}{\scriptsize Qwen2.5-VL-72B-Instruct} & Baseline & \multirow{6}{*}{24} & 87.0 & 43.6 & 37.1\\
     & w/o \(\delta_{T,l}^h\) &  & 89.8 & 45.3 & 40.3\\
     & w/o \(\delta_{V,l}^h\) &  & 89.0 & 50.9 & 49.2\\
     & Rnd-\(\Delta\) &  & 86.2 & 43.6 & 39.1\\
     & \(-\alpha\Delta\) &  & 79.7 & 35.5 & 25.8\\
     & \(+\alpha\Delta\) &  & \textbf{91.5} & \textbf{59.8} & \textbf{57.3}\\
    \bottomrule
  \end{tabular}
\end{table}

Table~\ref{table-large-model} shows that VisionToM remains effective on Qwen2.5-VL-72B. The intervention improves the 72B backbone on all three EgoToM tasks and surpasses strong large-model baselines on Belief and Actions, indicating that the method remains beneficial even when the base MLLM is already strong.

\subsection{Cross-Dataset Transfer on MMToM-QA}

\begin{table}[h]
  \centering
  \footnotesize
  \setlength{\tabcolsep}{0.6mm}
  \renewcommand{\arraystretch}{0.90}
  \caption{Experiments on the MMToM-QA benchmark under the video-only setting. ``Transfer'' directly applies the intervention vector learned on EgoToM without retraining on MMToM-QA.}
  \label{table-mmtom}
  \begin{tabular}{@{}cccccc@{}}
    \toprule
    Method & Setting & Nframe & Belief & Goal & All \\
    \midrule
    {\scriptsize BIP-ALM-LLaMA2} & \multirow{4}{*}{\scriptsize Video Only} & - & 64.0 & 58.3 & 61.2\\
    {\scriptsize Qwen2.5-VL-7B-Instruct} &  & \multirow{3}{*}{8} & 47.0 & 29.3 & 38.2\\
    \textbf{{\scriptsize VisionToM w/ Qwen2.5-VL-7B-Instruct}} &  &  & \textbf{70.7} & \textbf{62.0} & \textbf{66.3}\\
    {\scriptsize Qwen2.5-VL-7B-Instruct Transfer} &  &  & 64.3 & 56.7 & 60.5\\
    \bottomrule
  \end{tabular}
\end{table}

On MMToM-QA, we evaluate both in-domain generalization and zero-shot transfer under the video-only setting. Following the same protocol as EgoToM, we compute intervention vectors from the MMToM-QA training split and evaluate on its benchmark. VisionToM achieves the best overall performance, while directly transferring the intervention vector learned on EgoToM yields results close to the strongest video-only baseline. These findings suggest that the learned directions capture transferable ToM reasoning patterns rather than dataset-specific shortcuts.

\newpage
\section{Additional Probing Results}
\label{additionalProbing}

\begin{figure*}[h]
  \centering
  \includegraphics[width=\linewidth]{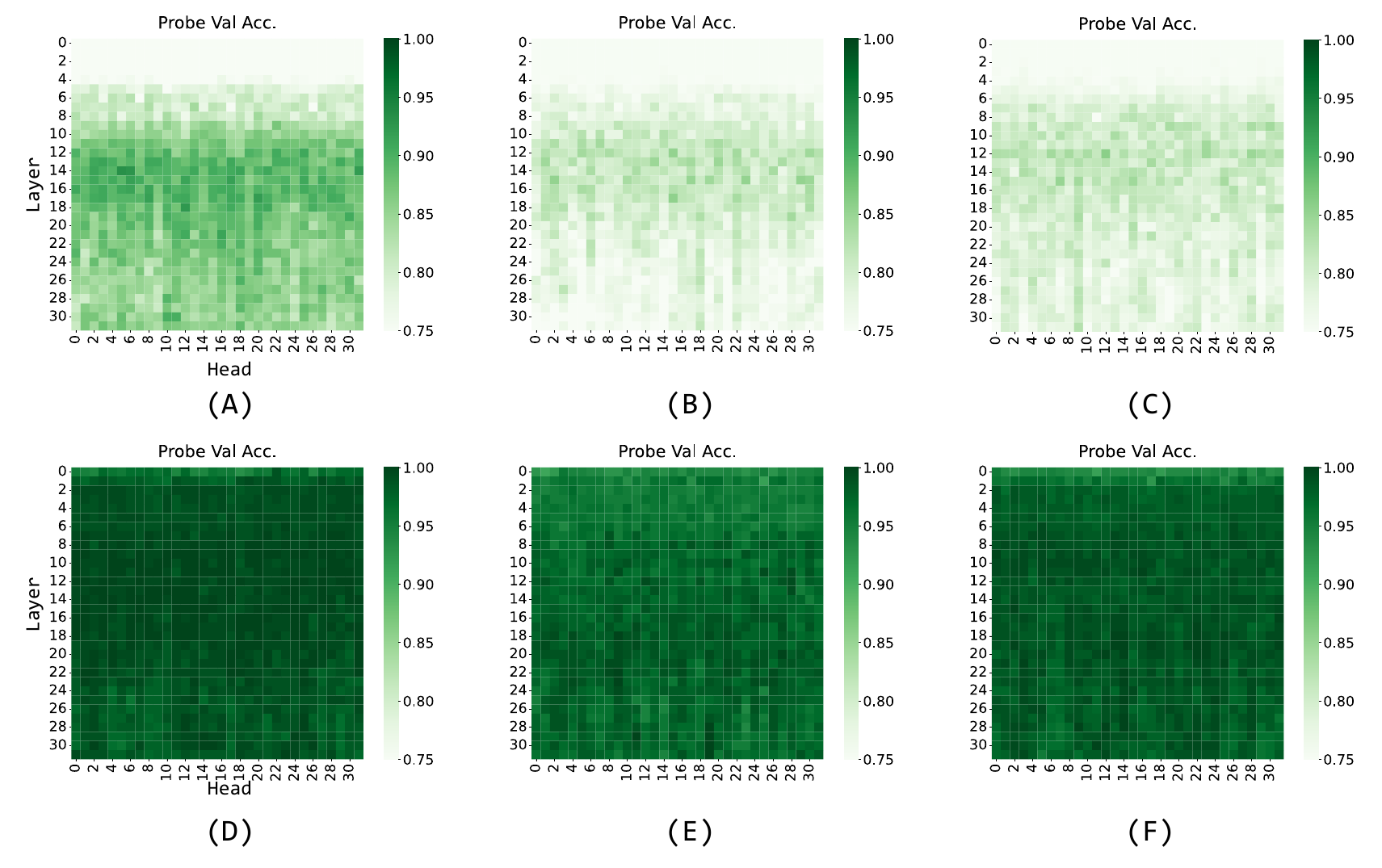}
  \caption{{
Probe validation accuracies for the three EgoToM tasks, based on activations from each attention head across all layers of LLaVA‑Next‑Video‑7B.
Subfigures (A)–(C) correspond to the ToM reasoning stage, showing accuracies for the (A) goal prediction, (B) belief inference, and (C) actions inference tasks, respectively.
Subfigures (D)–(F) correspond to the visual attention stage, showing the same tasks in the order: (D) goal prediction, (E) belief inference, and (F) actions inference.
Darker shades indicate higher probing accuracy, suggesting stronger task-relevant signals in specific heads and layers.}}
  \label{probingAll}
\end{figure*}

\begin{figure*}[h]
  \centering
  \includegraphics[width=\linewidth]{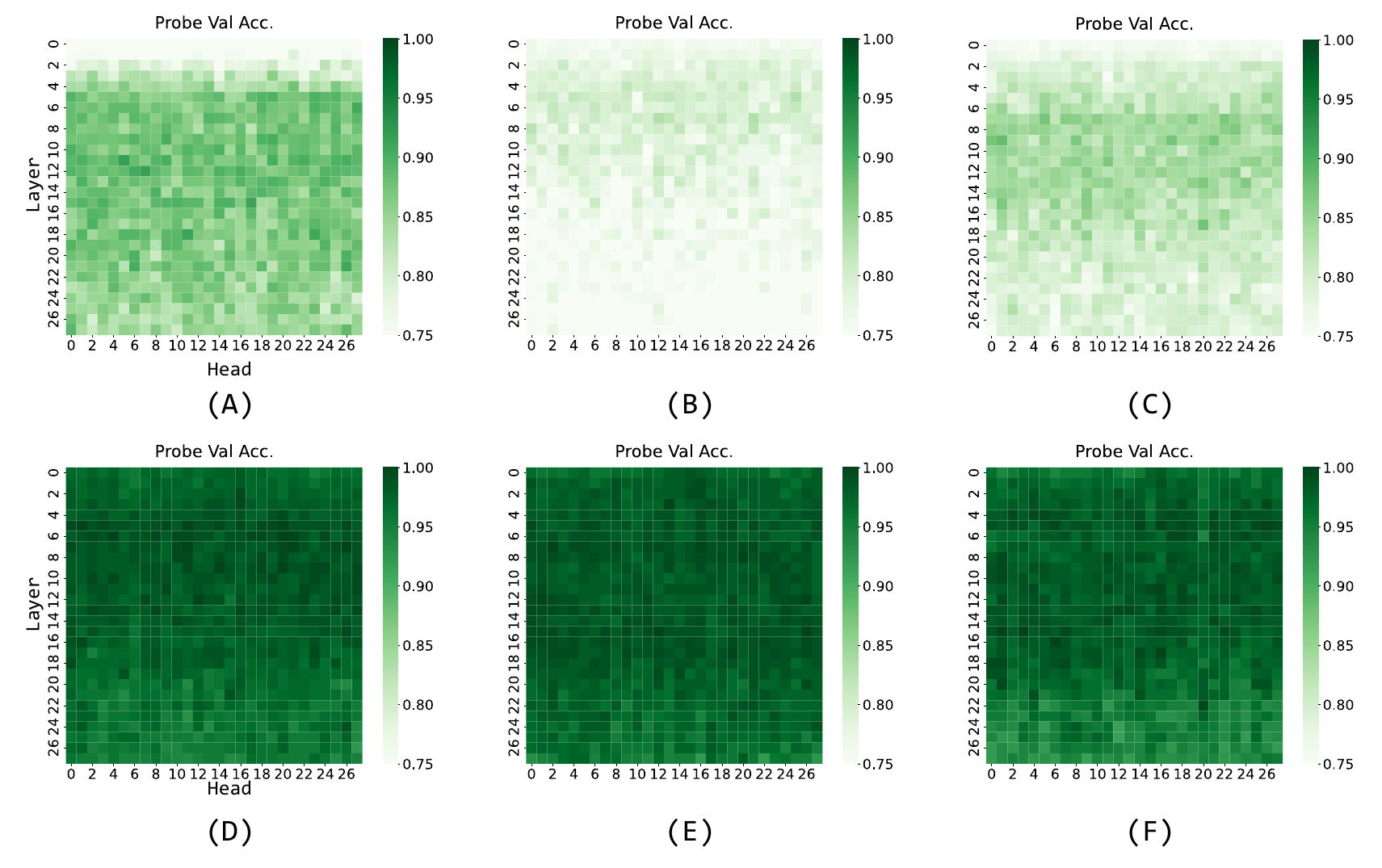}
  \caption{{Probe validation accuracies for the three EgoToM tasks, based on activations from each attention head across all layers of Qwen2.5-VL-7B.
Subfigures (A)–(C) correspond to the ToM reasoning stage, showing accuracies for the (A) goal prediction, (B) belief inference, and (C) actions inference tasks, respectively.
Subfigures (D)–(F) correspond to the visual attention stage, showing the same tasks in the order: (D) goal prediction, (E) belief inference, and (F) actions inference.
Darker shades indicate higher probing accuracy, suggesting stronger task-relevant signals in specific heads and layers.}}
  \label{probingAll-qwen}
\end{figure*}

The probing results on the LLaVA-Next-Video and Qwen2.5-VL models are shown in Figures~\ref{probingAll} and~\ref{probingAll-qwen}, covering two stages: visual attention probing and ToM reasoning probing. Each stage includes independent probing of Goal, Belief, and Actions tasks, with the y-axis representing attention layers and the x-axis representing attention heads.

\newpage
\section{Hyperparameters' Analysis}
\label{hyperparameters}

\begin{figure}[h]
  \centering
  \includegraphics[width=\linewidth]{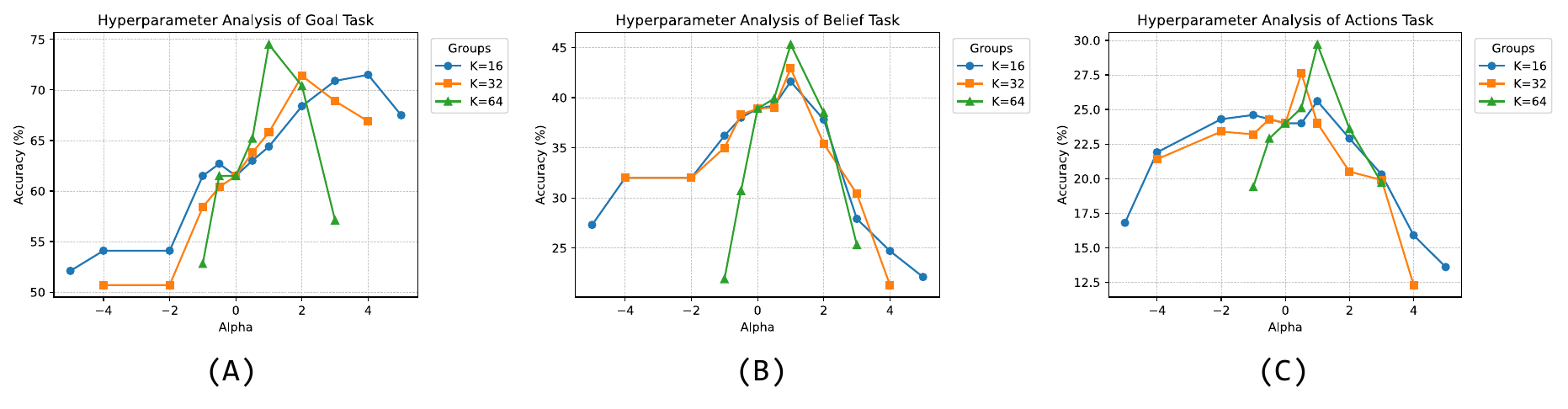}
  \caption{Analysis of the Impact of Hyperparameter of LLaVA-Next-Video on Three Tasks}
  \label{HyperparametersAnalysisLLaVA}
\end{figure}

\begin{figure}[h]
  \centering
  \includegraphics[width=\linewidth]{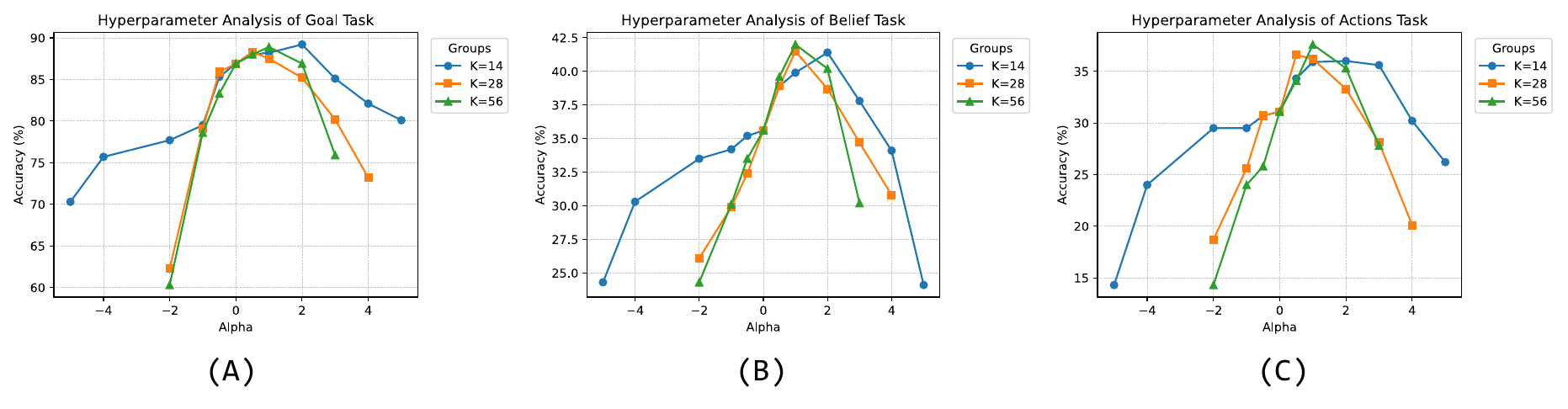}
  \caption{Analysis of the Impact of Hyperparameter of Qwen2.5-VL on Three Tasks}
  \label{HyperparametersAnalysisQwen}
\end{figure}

Figure~\ref{HyperparametersAnalysisLLaVA} and Figure~\ref{HyperparametersAnalysisQwen} respectively show the intervention effects of the number of editing heads \(K\) and intervention strength \(\alpha\) on the LLaVA-Next-Video model and Qwen2.5-VL model across three tasks in the EgoToM benchmark. The three subplots correspond to: (A) Goal Task, (B) Belief Task, and (C) Actions Task.

For the editing heads \(K\) of the LLaVA-Next-Video model, we choose to use 16, 32, 64 based on its attention head count of 32. For the Qwen2.5-VL model with an attention head count of 28, we choose to use 14, 28, 56 as the editing heads \(K\). We did not search for the optimal editing heads \(K\) to achieve the best results.

Theoretically, whether for visual attention enhancement or ToM reasoning guidance, the obtained \(\delta\) represents a correction from negative samples pointing to positive samples, so \(+\alpha\) intervention will bring positive gains, while \(-\alpha\) will weaken model capabilities. The results in Figure~\ref{HyperparametersAnalysisLLaVA} and~\ref{HyperparametersAnalysisQwen} strongly support our hypothesis, with intervention effects showing monotonic behavior around the baseline (\(\alpha=0\)) and presenting uniform and coherent characteristics within the effective range. Additionally, the improvement effects brought by VisionToM intervention are not unlimited, but are only effective within a certain intervention strength range (for the LLaVA-Next-Video model, this range is \(\alpha \in [-5, 5]\)). Beyond this range, all responses become invalid. Specifically, as shown in Figure~\ref{HyperparametersAnalysisLLaVA}(A), when \(\alpha\)=4, the accuracy rates for \(K\)=16, 32, and 64 all show a declining trend. Unlike the decline when \(\alpha\)=-1, the performance degradation here is mainly due to excessive interference intensity, causing some responses to become invalid (such as outputting garbled text or infinitely repeating words). During the statistical process, we retained all samples and treated invalid responses as errors to ensure consistency in comparison. The same phenomenon was observed in other experiments, indicating that our method has controllability and remains effective within a certain range of interference.

\newpage
\section{Open-ended Generation}
\label{openEndedQA}

We have listed examples of open-ended generation here, including the responses from the base model as well as the improved effects after applying our method.

\begin{figure}[!h]
  \centering
  \includegraphics[width=0.9\linewidth]{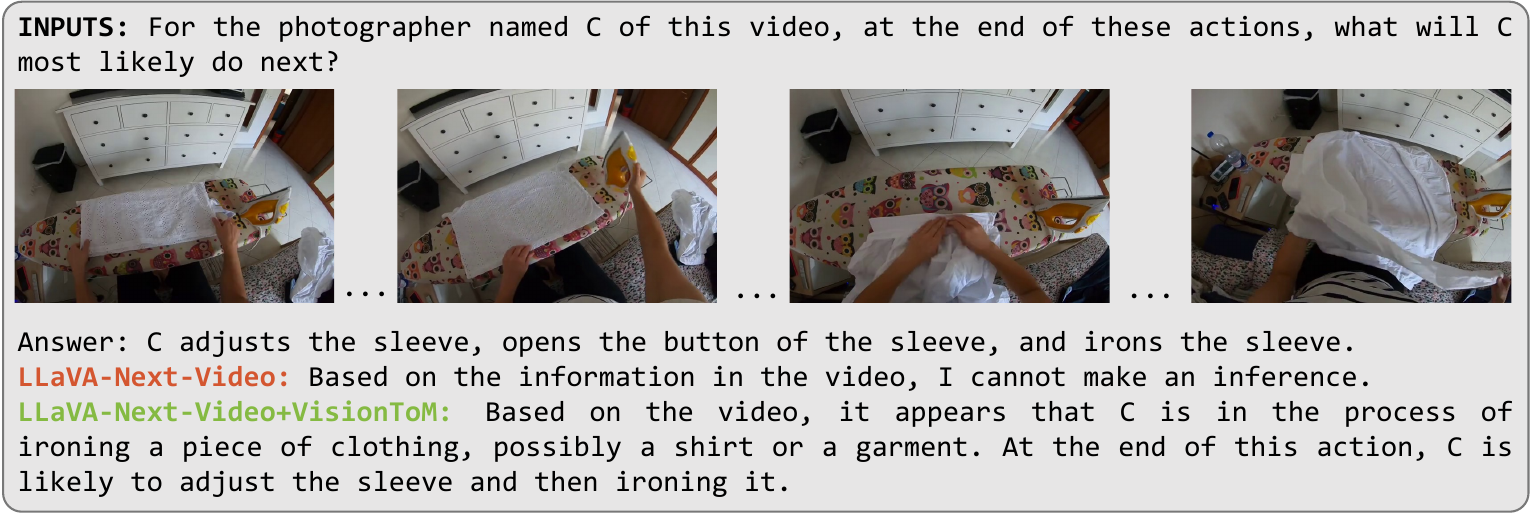}
\end{figure}

\begin{figure}[!h]
  \centering
  \includegraphics[width=0.9\linewidth]{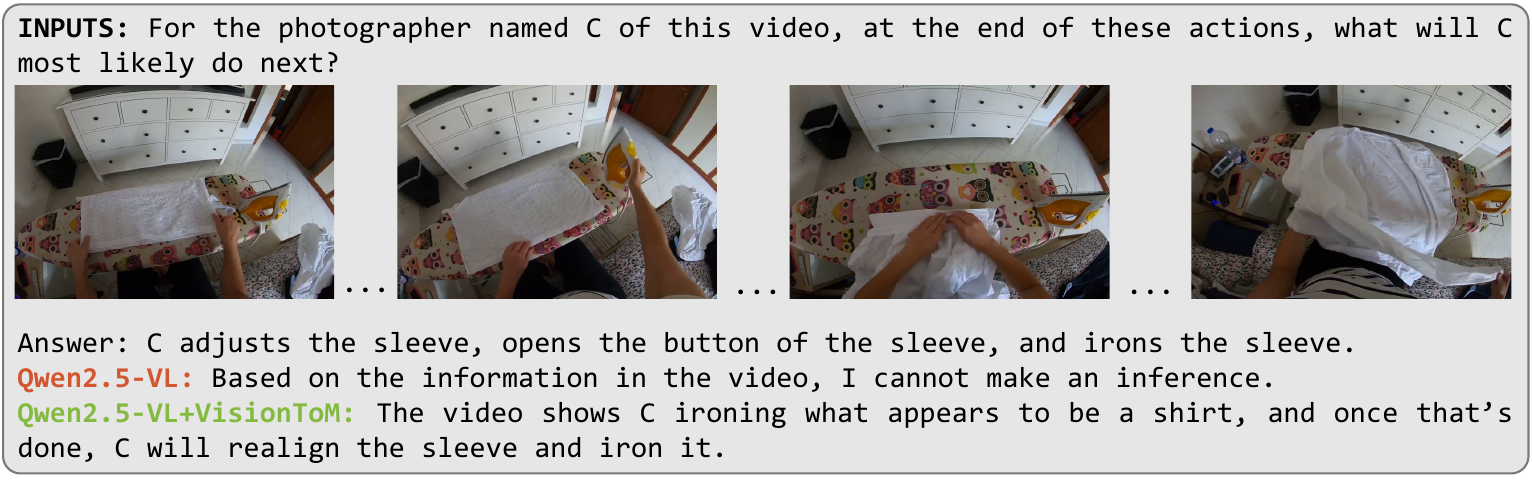}
\end{figure}

\begin{figure}[!h]
  \centering
  \includegraphics[width=0.9\linewidth]{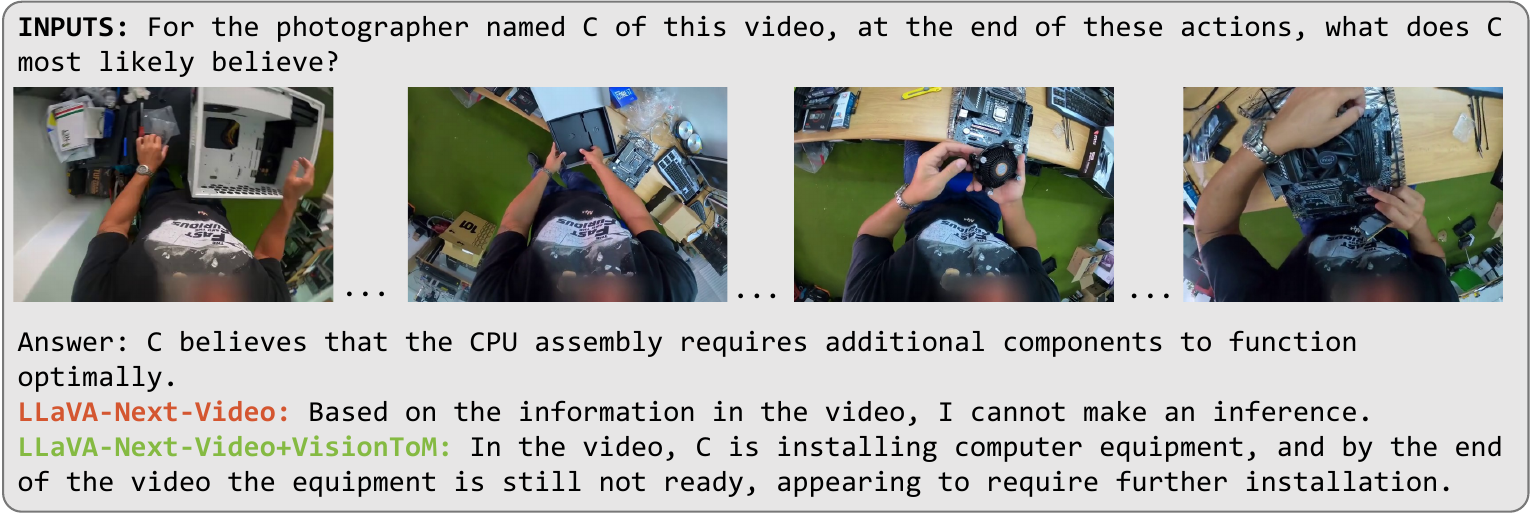}
\end{figure}

\begin{figure}[!h]
  \centering
  \includegraphics[width=0.9\linewidth]{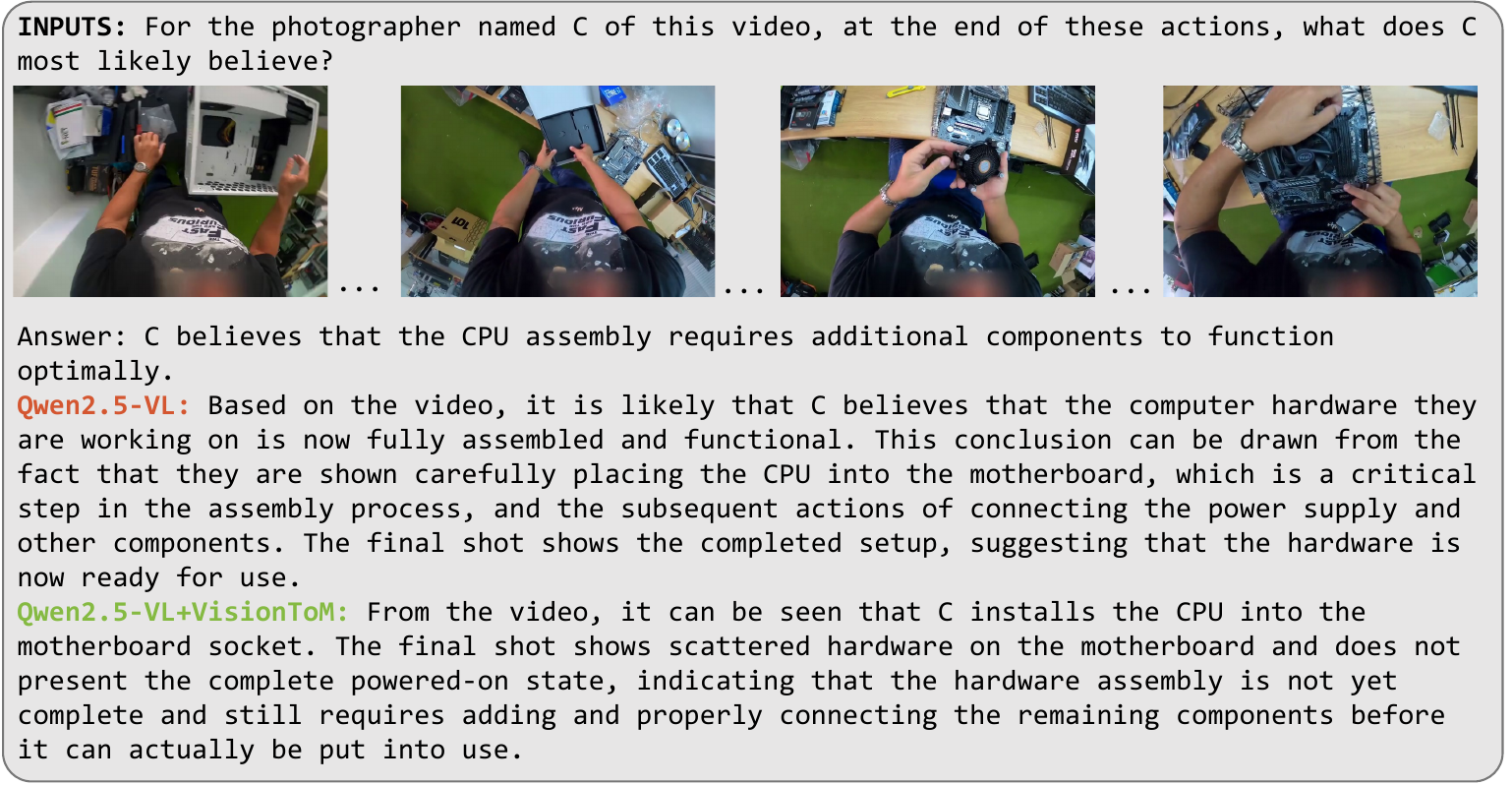}
\end{figure}

\begin{figure}[!h]
  \centering
  \includegraphics[width=0.9\linewidth]{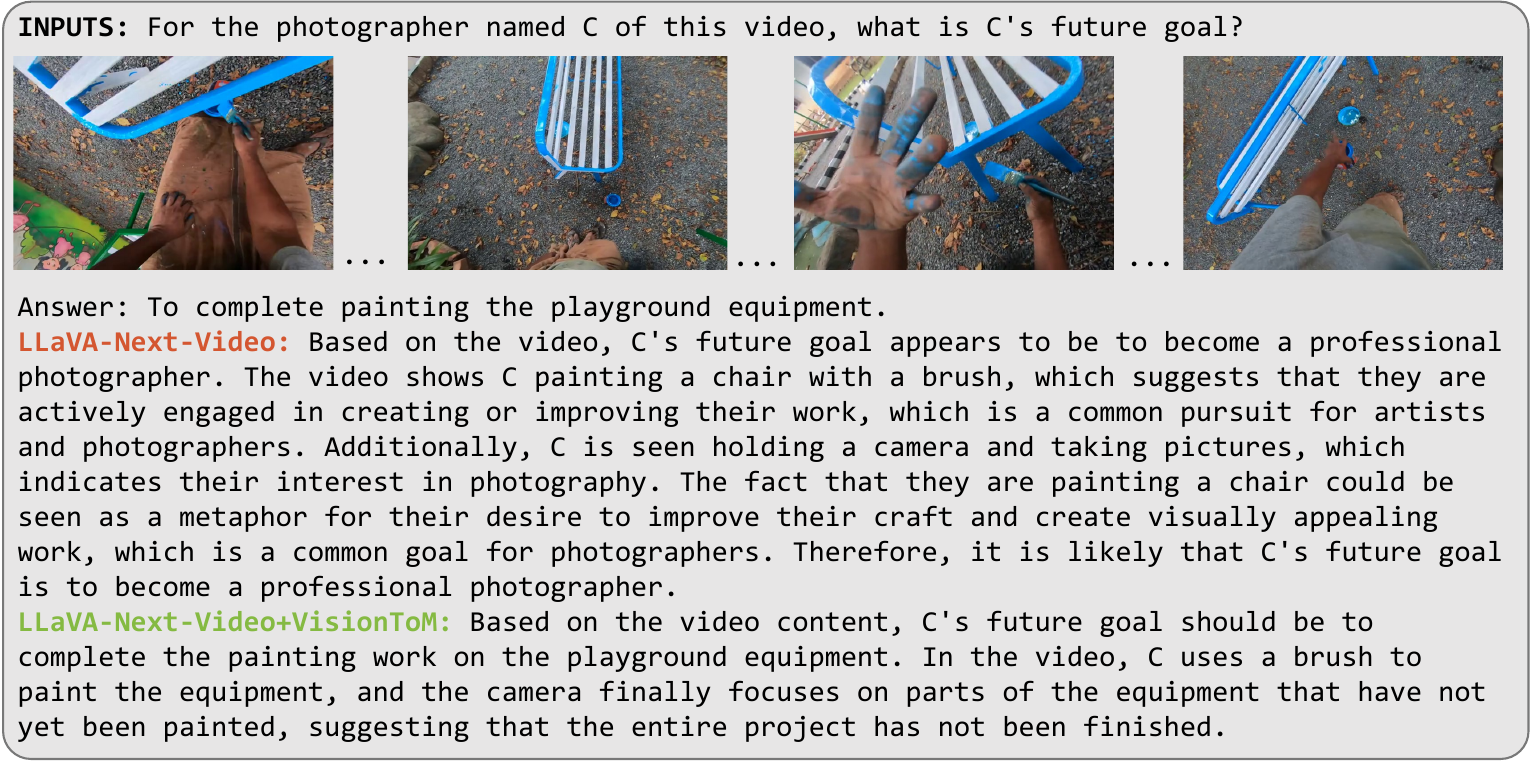}
\end{figure}

\begin{figure}[!h]
  \centering
  \includegraphics[width=0.9\linewidth]{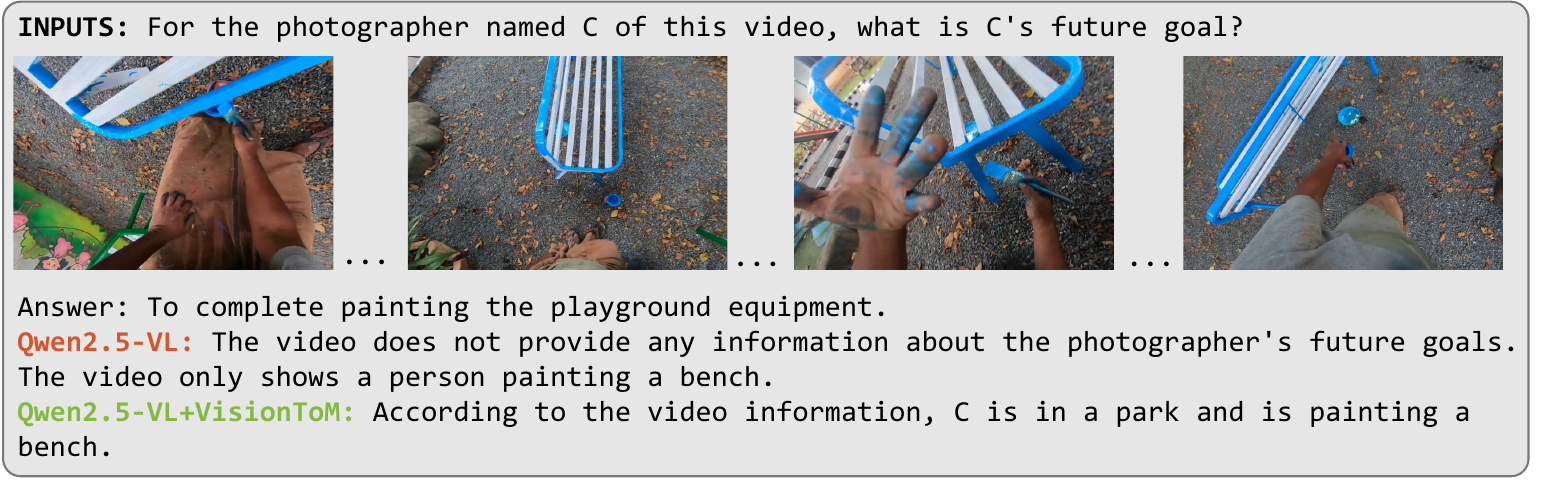}
\end{figure}

\newpage
\section{Experiments Settings}
\label{experimentsSettings}

\subsection{Data preprocessing}
\label{dataPreprocessing}

We follow the experimental setup of the EgoToM dataset, extracting corresponding video segments from the Ego4D dataset based on the timeline it provides. EgoToM includes three ToM tasks: Goal, Belief, and Actions, with sample sizes of 351, 335, and 354 respectively. According to the experimental protocol, we sample video frames at equal intervals from each video segment and input these frames along with the corresponding questions into the model as the sole source of information for reasoning. For each task, we use a 30\% calibration split to train the probe and encoder and compute intervention vectors, and a disjoint 70\% evaluation split for final testing. No labels or answers from the evaluation split are used when learning the intervention directions.

\subsection{Computing infrastructure}
\label{computingInfra}

To ensure reproducibility, all experiments were conducted under the following computing environment: Ubuntu 22.04; 14 vCPUs on an Intel® Xeon® Gold 6348 @ 2.60 GHz; 8$\times$NVIDIA A800 GPUs; and 100 GB system memory. The software stack consists of Python 3.12, PyTorch 2.5.1, and CUDA 12.4. We fixed the global random seed to 42 and enabled deterministic settings to eliminate randomness from data loading and operator-level execution. Both training and inference were performed in FP16 precision.

\subsection{Calibration Cost}

VisionToM keeps the MLLM backbone frozen. On the hardware reported in Section~\ref{computingInfra}, the one-time calibration stage takes approximately 0.2 hours for probe training and 1 hour for encoder training. All downstream experiments, including multiple-choice QA, open-ended generation, large-model evaluation, and MMToM-QA transfer, directly apply the resulting precomputed intervention vectors without further training.

\subsection{Open-ended Evaluation Details}

For each open-ended answer, two DeepSeek-R1 judges are prompted independently, and we accept a label only when both judges agree. The prompt explicitly defines the ``true'' and ``info'' criteria and standardizes edge cases. In particular, an answer is marked ``false'' if any factual statement is incorrect, hallucinated, logically contradictory, or inconsistent with the reference facts; answers that mix correct and incorrect claims are also marked ``false''. An answer is marked ``info'' only if it contains substantive task-relevant content rather than vague restatements. We additionally performed manual verification with three volunteers and observed human--LLM agreement rates of 96.2\% for the ``true'' label and 93.5\% for the ``info'' label.